\documentclass[runningheads]{llncs}
\usepackage{amsmath,graphicx}
\usepackage[english]{babel}
\usepackage[utf8]{inputenc}
\usepackage{algorithm}
\usepackage{algorithmic}
\usepackage{graphicx}
\usepackage{textcomp}
\usepackage{xcolor}
\usepackage{booktabs}
\usepackage{array}
\usepackage{bm}
\usepackage{bbm}
\usepackage{float}
\usepackage{stfloats}
\usepackage{wrapfig}
\usepackage{microtype}
\usepackage{graphicx}
\usepackage{subfigure}
\usepackage{xspace}
\usepackage{amsmath}
\usepackage{amssymb}
\usepackage{amsfonts}
\usepackage{bm}
\usepackage{bbm}
\usepackage{float}
\usepackage{stfloats}
\usepackage{wrapfig}
\usepackage[T1]{fontenc}
\usepackage{lmodern}
\usepackage{multirow}
\usepackage{lipsum}
\usepackage{mathtools}
\usepackage{cuted}
\usepackage{stfloats}
\usepackage{multirow}
\usepackage{algorithm}
\usepackage{algorithmic}
\usepackage{amsmath}
\DeclareMathOperator*{\argmax}{arg\,max}
\DeclareMathOperator*{\argmin}{arg\,min}
\usepackage[font=small]{caption}

\begin{document}
%
% \title{Towards Robust Robot Perception: Enhancing Offline Reinforcement Learning with Adversarial Attacks and Defenses\thanks{Supported by organization x.}}
\title{Towards Robust Policy: Enhancing Offline Reinforcement Learning with Adversarial Attacks and Defenses}
\author{Thanh Nguyen*\inst{1} \and
Tung M. Luu*\inst{1} \and
Tri Ton\inst{1} \and
Chang D. Yoo\inst{1}}
\authorrunning{Thanh Nguyen, Tung M. Luu et al.}

\institute{Korea Advanced Institute of Science and Technology (KAIST), Korea\\
* Equal contribution\\
\email{thanhnguyen@kaist.ac.kr}\\
\url{https://www.kaist.ac.kr/}}

% \institute{Korea Advanced Institute of Science and Technology (KAIST), Korea \and \email{\{thanhnguyen,tungluu2203,tritth,cd_yoo\}@kaist.ac.kr}}
% %
\maketitle              % typeset the header of the contribution
\begin{abstract}
Offline reinforcement learning (RL) addresses the challenge of expensive and high-risk data exploration inherent in RL by pre-training policies on vast amounts of offline data, enabling direct deployment or fine-tuning in real-world environments. However, this training paradigm can compromise policy robustness, leading to degraded performance in practical conditions due to observation perturbations or intentional attacks. While adversarial attacks and defenses have been extensively studied in deep learning, their application in offline RL is limited. This paper proposes a framework to enhance the robustness of offline RL models by leveraging advanced adversarial attacks and defenses. The framework attacks the actor and critic components by perturbing observations during training and using adversarial defenses as regularization to enhance the learned policy. Four attacks and two defenses are introduced and evaluated on the D4RL benchmark. The results show the vulnerability of both the actor and critic to attacks and the effectiveness of the defenses in improving policy robustness. This framework holds promise for enhancing the reliability of offline RL models in practical scenarios.

\end{abstract}
\section{Introduction}

Recent advancements in Deep Neural Networks (DNN) \cite{krizhevsky2017imagenet,he2016deep,devlin2018bert,vaswani2017attention,vu2022softgroup,vu2023scalable,nguyen2023dimcl} have greatly influenced the success of Deep Reinforcement Learning (RL) \cite{levine2016end,schulman2015trust,mnih2015human,silver2016mastering}
\cite{nguyen2021sample,nguyen2021robust}, addressing the challenge of high dimensionality. However, traditional RL data collection methods rely on online interactions with the environment, which can be costly and pose safety concerns in practical applications like robotics and healthcare \cite{levine2020offline}. To overcome these limitations, Offline RL has emerged as a promising alternative, enabling policy learning from pre-collected data without online interaction. However, non-interactive training in offline RL introduces challenges related to distributional shift and robustness \cite{levine2020offline,fujimoto2019off,kumar2019stabilizing}. While numerous recent methods in offline RL have been developed to tackle distributional shift  \cite{chen2019information,jin2021pessimism,rashidinejad2021bridging,xie2021bellman,yin2022near,fujimoto2019off,kumar2019stabilizing,kumar2020conservative}, there has been scant attention directed towards enhancing robustness, particularly in mitigating performance degradation under real-world conditions such as observation perturbations stemming from sensor errors or intentional attacks \cite{zhang2022corruption,yang2022rorl}. Hence, there is a pressing need to introduce novel techniques aimed at bolstering the robustness of offline RL methods.

To this end, this paper introduces a framework to enhance the robustness of learned models in offline RL by leveraging advanced adversarial attacks and defenses. More specifically, our framework entails introducing perturbations to observations during training, strategically targeting both the actor and critic components of offline RL. Simultaneously, we incorporate adversarial defenses as regularization techniques. The overarching goal of this approach is to proactively train the policy to fend off potential attacks, ultimately fortifying its robustness when deployed in real-world scenarios. In our study, we put forth targeted attacks on the actor and critic components, encompassing the Random Attack, Critic Attack, Robust Critic Attack, and Actor Attack. These attacks shed light on the vulnerability of actor and critic components in offline RL methods to observation perturbations, demonstrated by the significant performance degradation they induce. To mitigate these vulnerabilities, we introduce adversarial defenses in the form of regularizers during the training process, specifically identified as Critic Defense and Actor Defense. We empirically prove that these defense mechanisms are effective in enhancing the robustness of model-free offline RL algorithms.

We conducted evaluations of these techniques using the D4RL benchmark \cite{fu2020d4rl}. Remarkably, the Robust Critic Attack emerged as the most potent, followed by the Critic Attack, Actor Attack, and Random Attack. Adversarial attacks were found to substantially undermine the performance of offline RL methods. Importantly, the incorporation of our proposed defenses within the training framework consistently resulted in improved learned policy performance across various datasets, in both clean and attacked environments.

The paper is structured as follows: Section 1 provides an introduction, followed by Section 2 discussing related work. Section 3 presents background information on offline RL, and Section 4 details the methodology, including the offline RL framework, proposed adversarial attacks, and adversarial defenses. Section 5 presents the experimental setup and results, and finally, Section 6 concludes the paper and discusses future directions.

\section{Related work}

\textbf{Offline RL.} This paper focuses on improving the robustness of model-free offline reinforcement learning (RL). \textit{Offline RL} is a research branch of RL that learns policies through a static dataset as opposed to online interaction of conventional \textit{online RL} \cite{levine2020offline,luu2022visual,luu2022utilizing}. In the model-free domain, offline RL methods usually attempt to correct the extrapolation error \cite{fujimoto2019off} in off-policy algorithms. A line of work focuses on regularizing the learned policy near the dataset distribution \cite{wu2019behavior,kumar2019stabilizing,fujimoto2019off,kumar2020conservative,fujimoto2021minimalist,kostrikov2022offline,bai2022pessimistic}. For instance, BCQ \cite{wu2019behavior} uses a direct policy constraint to force the learned policy to be close to the behavior policy, estimated by using a parametric generative model. Alternatively, BEAR \cite{kumar2019stabilizing} proposes using support matching (e.g., maximum mean discrepancy (MMD) divergence) to constrain the learned policy, showing better performance. Another line of work mitigates the selection of out-of-distribution (OOD) actions during policy evaluation by penalizing the value function \cite{kumar2020conservative,agarwal2020optimistic,an2021uncertainty,bai2022pessimistic,nguyen2023fast}. For example, Conservative Q-Learning (CQL) \cite{kumar2020conservative} proposes the regularization while learning the $Q$-value function  by pushing up values for state-action pairs seen in the dataset and pulling down values in unseen actions. Notably, TD3+BC \cite{fujimoto2021minimalist} simply uses behavioral cloning to constrain the learned policy on top of the conventional TD3 \cite{fujimoto2018addressing} but achieves competitive performance with other complex offline RL algorithms. In contrast, EDAC \cite{an2021uncertainty} leverages ensemble $Q$ networks and proposes an objective to diversify their gradients to prevent overestimation when learning the $Q$-value function. Similarly, PBRL \cite{bai2022pessimistic} proposes explicit value underestimation of OOD actions according to uncertainty, requiring fewer ensemble networks.

\textbf{Robust RL against Adversarial Attacks on State Observations.} In the \textit{online RL} setting, where the algorithm is allowed to interact with the environment during training, the vulnerabilities in state observations caused by adversarial attacks were first demonstrated in \cite{huang2017adversarial,lin2017tactics,kos2017delving}. In the continuous control domain, \cite{pattanaik2018robust} introduced an attack that exploits both the policy and the $Q$-value function to craft the perturbations. Recent work by \cite{zhang2020robust} formalized attacks on observations through a state-adversarial MDP (SA-MDP), demonstrating that the most powerful attacks can be learned as an RL problem. Building on this, \cite{zhang2021robust} and \cite{sun2021strongest} introduced RL-based attackers for black-box and white-box attacks, respectively. To enhance the robustness of online RL agents against adversarial attacks on observations, several methods have been proposed. One family of methods aims to improve the robustness of deep neural network (DNN) components in RL algorithms by enforcing properties such as invariance and smoothness under bounded perturbations. For instance, \cite{shen2020deep,zhang2020robust} introduced regularizers to encourage the smoothness of the policy network under adversarial perturbations. More recently, another group of methods has focused on training the agent against a learnable attacker. Typically, \cite{zhang2021robust,sun2021strongest} proposed concurrently training the agent together with an RL-based attacker, leading to a more robust RL agent. 
% Although training with an RL-based attacker has shown promising results, it is only applicable in the online setting due to the requirement for online training of the attacker.
Overall, most of the mentioned methods are focusing on online RL where policies can collect more data in the environment to correct the attacking states. 
In the \textit{offline RL} setting, insufficient attention to robustness is observed \cite{zhang2022corruption}. An example of research addressing this concern is RORL \cite{yang2022rorl}, which employs adversarial defense strategies to enhance model robustness. RORL is grounded in an uncertainty-based offline RL algorithm, relying on an ensemble of actors or critics.
Offline RL, in which agents are trained on a static dataset, exhibits distinct characteristics compared to online RL. The question of whether we can effectively adapt online RL methods to offline RL remains uncertain. 
% Therefore, the robustness of offline RL algorithms remains a question. 

In contrast, our investigation thoroughly explores the robustness of well-established offline RL algorithms, including BCQ \cite{wu2019behavior}, CQL \cite{kumar2020conservative}, and TD3+BC \cite{fujimoto2021minimalist}, with the aim of identifying the most versatile and effective robust approaches. Our objective is to establish a comprehensive baseline for both attack and defense scenarios for reference. By doing so, we aspire to provide valuable insights that can guide future endeavors dedicated to enhancing the overall robustness of offline RL.

\section{Background}
Mathematically, offline RL addresses the challenge of learning to control a dynamic system, which can be defined as a Markov decision process (MDP). The MDP is defined by a tuple $M = (S,A,T,d_0,r,\gamma)$. Therein, $S$ is a set of state $s$, $A$ is a set of actions $a$, $T$ is the transition probability of the dynamics in the form $T(s_{t+1}|s_t,a_t)$, $d_0$ is the initial state distribution, $r$ is reward function, $\gamma \in (0,1)$ is a scalar discount factor and H is the horizon. 

Within a MDP, there is a policy $\pi(a_t|s_t)$ for controlling the dynamic. A trajectory distribution, which is a sequence of $H+1$ states and $H$ actions, can be further derived as $\tau = (s_0,a_0,...,s_H)$ where H can be infinite. The probability density function for a given trajectory $\tau$ under policy $\pi$ is as below:
\begin{align}
p_{\pi}(\tau)=d_{0}\left(\mathbf{s}_{0}\right) \prod_{t=0}^{H-1} \pi\left(\mathbf{a}_{t} \mid \mathbf{s}_{t}\right) T\left(\mathbf{s}_{t+1} \mid \mathbf{s}_{t}, \mathbf{a}_{t}\right)   
\end{align}
Given a static dataset $D=\{(s^i_t,a^i_t,s^i_{t+1},r^i_t)\}$ collected by an unknown behavioural policy $\pi^\beta$, offline RL tries to learn an optimal policy $\pi^*$ that maximizes the expected return. Mathematically, this involves solving the return maximization problem:
\begin{equation}
\pi^{*}=\underset{\pi}{\operatorname{argmax}} \quad \mathbb{E}_{\tau \sim p_{\pi}}\left[\Sigma_{t=0}^{H-1} \gamma^{t} r\left(s_{t}, a_{t}\right)\right].
\end{equation}

To aid in solving MDP, it is common practice to define the value function $V^\pi(s)=\underset{\tau \sim \pi}{\mathbb{E}}\left[R(\tau) \mid s_0=s\right]$ where $R(\tau)=\sum_{t=0}^{H} \gamma^t r_t$ is the discounted return. This function provides the expected return when initiating from state $s$ and consistently following policy $\pi$. Additionally, the action value function $Q^\pi(s, a)=\underset{\tau \sim \pi}{\mathbb{E}}\left[R(\tau) \mid s_0=s, a_0=a\right]$ is defined, representing the expected return when beginning from state $s$, taking action $a$, and subsequently adhering to policy $\pi$.

\begin{algorithm}[htpb]
	\caption{Generic model-free offline RL}
	\label{alg:generic_algo}
	\begin{algorithmic}[1]
	\STATE \textbf{Input}: Max iterations K, Batch size E, Offline dataset D, optionally estimated behavior $\hat{\pi}^\beta$\\
        Initialize $\pi_0$ and $Q^{\pi_0}$ randomly.
        \FOR {$ k = 1 \text{ \textbf{to} } K $}
	\STATE Sampling from replay buffer:
        \begin{equation*}
            B =  \{(s_j, a_j, s_{j+1}, r_j)\}_{j=1}^E \sim D
        \end{equation*}

	\STATE Policy evaluation: $Q^{\pi_{k-1}}=\mathcal{T}\left(\pi_{k-1}, B, Q^{\pi_{k-2}}\right)$ 
         
	\STATE Policy improvement: $\pi_{k}=\mathcal{V}\left(Q^{\pi_{k-1}}, \hat{\pi}^\beta, B, \pi_{k-1}\right)$
		
        \ENDFOR
	\end{algorithmic}
\end{algorithm}
In solving MDP, existing literature on offline RL frequently employs interactive actor-critic approaches \cite{fujimoto2019off,kumar2019stabilizing,kumar2020conservative}.  The generic model-free offline RL algorithms can be summarized as approximate modified policy iteration \cite{brandfonbrener2021offline} as depicted in the \textbf{Algorithm} \ref{alg:generic_algo}. To be more specific, the algorithm is performed by iteratively learning the critic and the actor on the data sample from a given static dataset until they converge to the final solution. In the iterative process, the critic $Q^{\pi_{k-1}}$ is learned with \textit{policy evaluation} given the current policy $\pi_{k-1}$, mini-batch data $B$, and the reference previous policy $Q^{\pi_{k-2}}$. The actor $\pi_k$ is learn by the \textit{policy improvement} given the estimated $Q^{\pi_{k-1}}$, mini-batch data $B$, reference previous policy $\pi_{k-1}$, and an optional explicitly estimated behavior policy $\hat{\pi}^\beta$ \cite{brandfonbrener2021offline}.

% There are many ways to perform policy evaluation and policy improvement as denoted by the corresponding operator $\mathcal{T}$, $\mathcal{V}$ respectively.  For the policy evaluation operator $\mathcal{T}$, common techniques include fitted Q evaluation, TD-style learning with targeted networks, and Q ensembles for uncertainty measurement \cite{fujimoto2019off,kumar2019stabilizing,kumar2020conservative,fujimoto2018addressing}. Other operators, such as importance weighting \cite{munos2016safe} or pessimism \cite{kumar2020conservative}, can also be considered. For the policy improvement operator $\mathcal{V}$, constraints are often used and can be categorized into behavior cloning, constrained policy updates, regularized policy updates, and variants of imitation learning, which encourage policies to imitate the behavior policy or stay within its support \cite{fujimoto2019off,brandfonbrener2022incorporating,wang2020critic,chen2020bail}.

Various methods exist for the \textit{policy evaluation} and the \textit{policy improvement}, each associated with specific operators denoted as $\mathcal{T}$ and $\mathcal{V}$, respectively. For the policy evaluation operator $\mathcal{T}$, common techniques encompass fitted Q evaluation \cite{fujimoto2019off,kumar2019stabilizing,kumar2020conservative}, TD-style learning employing targeted networks \cite{fujimoto2018addressing}, and the utilization of Q ensembles for uncertainty measurement \cite{yang2022rorl}. Other operators, such as importance weighting \cite{munos2016safe} or pessimism \cite{kumar2020conservative}, are also viable options.

Concerning the policy improvement operator $\mathcal{V}$, constraints often play a pivotal role and can be categorized into various strategies. These include behavior cloning \cite{torabi2018behavioral}, constrained policy updates \cite{fujimoto2019off} regularized policy updates \cite{brandfonbrener2022incorporating}, and different forms of imitation learning \cite{wang2020critic,chen2020bail}. These strategies are designed to encourage policies to either imitate the behavior policy or remain within its support . The diversity in these approaches reflects the rich landscape of methods available for both policy evaluation and improvement in the realm of reinforcement learning.

Despite the variations among the aforementioned methods, a common thread unites many successful modern approaches—they heavily rely on deep neural networks (DNNs) to model policies, Q functions, or value functions. However, this reliance on DNNs becomes a vulnerability, as these networks are known to be susceptible to input perturbations \cite{huang2017adversarial,kos2017delving}. This susceptibility opens the door for attackers to manipulate the actor, the critic, or both simultaneously through observation perturbation. To illustrate, attackers can introduce a small amount of noise to the input observations of the actor, causing it to behave significantly differently compared to its original behavior and resulting in a trajectory with very low returns. Moreover, in certain scenarios, attackers may have access to the critic, allowing them to identify the suboptimal states of the actor and guide the actor towards these states, ultimately lowering the overall return.

% Regardless, the successful modern methods share a commonality as they relies heavily on deep neural networks (DNN) to model the  a policy, a Q function, or a value function . As a result, it becomes the source of attack since DNN is known to be vulnerable under input perturbation. That means attackers can attack the actor or the critic or both of them at the same time by observation perturbation. For instance, the attackers can inject a small amount of controllable noise to input observation of the actor to make it behave significantly different compared to the original one leading to a very low-return trajectory. Furthermore, in some settings, the attackers can be given access to the critic. This lets the attackers know the poor states of the actor and orientate the actor into them, aiming lower the return. 

\section{Methodology}

In this session, we will introduce the framework for robust offline Reinforcement Learning (RL) training. Subsequently, we will introduce four adversarial attacks and two defenses, carefully chosen to be incorporated into the framework.

\subsection{The framework for robust offline RL training}
This study strengthens conventional offline RL methods by introducing a modified training process that incorporates insights from adversarial attacks and defenses. The motivation 
is to mitigate the vulnerability of DNNs in RL systems to input perturbations, exploited by attackers to manipulate actor and critic networks. The framework integrates adversarial examples into training to expose the network to perturbations, fostering the learning of resilient features. By training on a mix of original and adversarially perturbed data, the networks improve generalization and exhibit robust behavior against attacks, thus enhancing overall performance and security.

More specifically, our framework augments conventional offline RL training objectives with an additional regularizer, or defense, to immunize models against adversarial examples. The training procedure closely follows the established process of generic offline RL with minimal modifications. Throughout the training phase, adversarial examples are generated by attacking the clean examples sampled from the replay buffer using a selected adversarial attack. These adversarial examples are then added with clean examples to train the model. The defense seamlessly integrates with either the 'policy evaluation' or 'policy improvement' objectives, contributing to robust regularization. This proactive training strategy ensures the model's proficiency in handling adversarial perturbations, thereby enhancing its overall robustness. Moreover, this design maintains comparability with the majority of generic model-free offline RL approaches, as depicted in \textbf{Algorithm} \ref{alg:generic_algo}. During the evaluation at test time, the trained models undergo scrutiny against a set of targeted attacks, which may align with or deviate from the initially selected training attack. This evaluation further validates the robustness of model with unseen attacks.

\subsection{Adversarial attack for offline RL}

We introduce four adversarial attacks aimed at undermining the performance of a trained RL agent in a testing environment. These attacks involve injecting small amounts of noise into the observations, causing the policy to generate suboptimal actions or low-return trajectories. Mathematically, we denote the attacker as $A$. To prevent unrestricted perturbations which is impractical, we limit the possible perturbation to an $\epsilon$-ball around the input observation. The adversarial example, i.e., the intentionally perturbed observation to cause a model mistake, is generated as $\tilde{s} = A(s, \epsilon)$, where $\tilde{s} \in B(s, \epsilon)$. $B(s, \epsilon)$ represents the set of adversary perturbations, containing all allowed $\epsilon$-neighbor perturbations of $s$.

The four proposed adversarial attacks are designed to specifically deceive the actor and critic, which are fundamental components of an offline RL method. The attack $A(.)$ can be one of the below attacks.

\textbf{Random Attack:} A simple method to attack is injecting random noise into the current state and hoping that the noise will cause the agent to take worse action.  This attack is referred to as \textit{Random Attack} and $A(.)$ is defined as follows:
\begin{equation}
\label{eq:random_attack (RAN)} \tilde{s} = s + \mathcal{N}(0, \epsilon)    
\end{equation}
where $\mathcal{N}(0, \epsilon)$ is a noise sampled from the Normal distribution with mean 0 and standard deviation $\epsilon$. 

\textbf{Critic Attack:} A smarter approach is utilizing the information from the critic
to generate a perturbation observation that leads the policy
to choose the worst possible action which gives the lowest
Q-value for the given observation. Since Q-value indicates
the estimation of return according to the action in a particular
state, choosing the lowest Q-value action for each state has a
high potential to get the lowest return trajectory which is the goal of the attack. This attack is referred to as \textit{Critic Attack} and $A(.)$ is defined as follows:
\begin{equation}
 \tilde{s} = \argmin_{\tilde{s} \in B(s,\epsilon)} Q_{\phi}(s, \pi_{\theta}(\tilde{s})). 
\end{equation}

\textbf{Robust Critic Attack:} The critic attack may not work well if the Q function is not correct enough. To be more specific, the critic attack relies heavily on the information of the Q function. If the Q-function is a perfect one, $\tilde{s} $ will leads to the worst action that minimizes the Q-value as expected. However, in practice, the Q function in offline RL is not a trust-able one, which may raise a drawback that if Q-function is poorly learned, the attack will fail to predict the correct perturbation since it relies heavily on Q information. This may affects the correctness of perturbation.

The \textit{Robust Critic Attack} represents an enhancement over the Critic Attack methodology by leveraging a high-quality Q-value function denoted as $Q_R^\pi(s,a)$. This specific Q-value function is obtained by training on a examination data buffer $\mathcal{R}$ collected through the policy, which we want to attack, on the test environment. It is essential to highlight that we assume the attack can examine the policy  in the beginning of testing phase with only a small interaction budget in the test environment then use the data to learn their own Q-function and then begin attack. Utilizing this acquired Q-value for the attack leads to a more accurate and robust perturbation generation. The loss function governing the learning process for $Q_R^\pi(s,a)$ is articulated as follows:

\begin{equation}
\begin{aligned}
    L_{R}&=\mathbb{E}_{(s,a,s',r) \sim \mathcal{R}} \left[ r+\gamma Q_{R}^\pi\left(s^{\prime}, a^{\prime}\right)-Q_{R}^\pi\left(s, a\right)\right]^2\\
    &+\lambda \mathbb{E}_{(s,a) \sim \mathcal{R}} \left[ \max _{\hat{a} \in B\left(a,\epsilon \right)}\left(Q_{R}^\pi\left(s, \hat{a}\right)-Q_{R}^\pi\left(s, a\right)\right)^2 \right].
\end{aligned}    
\end{equation}

The \textit{Robust Critic Attack} is then defined as:
\begin{equation}
  \tilde{s} = \argmin_{\tilde{s} \in B(s,\epsilon)} Q_R(s, \pi_{\theta}(\tilde{s})).
\end{equation}

\textbf{Actor Attack:} In scenarios where only access to the actor is allowed and the critic information is not available, we propose \textit{Actor Attack}. Inspired by \cite{zhang2020robust}, this attack find the adversarial example maximizes the KL-divergence between output of actor with clean example and output of actor with corresponding adversarial example. Mathematically, it is defined as:

\begin{equation}
    \label{eq:actor_mad_attack}
        \tilde{s} = \argmax_{\tilde{s} \in B(s,\epsilon)} KL(\pi(.|s) || \pi(.|\tilde{s})).
\end{equation}

For actions parameterized by Gaussian mean $\pi_{\theta}(s)$ and covariance matrix $\Sigma$ (independence of $s$), the equation for A(.) becomes:
\begin{equation}
    \tilde{s} = \argmax_{\tilde{s} \in B(s,\epsilon)} (\pi_{\theta}(s)-\pi_{\theta}(\tilde{s}))^{\top}\Sigma^{-1}(\pi_{\theta}(s)-\pi_{\theta}(\tilde{s})).
\end{equation}

\subsection{Adversarial defenses for offline reinforcement learning}

When presented with adversarial examples generated by any of the aforementioned adversarial attacks, the policy undergoes training on these examples following the standard offline Reinforcement Learning (RL) protocol, augmented with an additional regularizer, i.e., defense. We introduce two defense for improving the robustness of the critic and actor, referred to as "Critic defense" and "Actor defense" respectively.

\textbf{Critic defense:} It is possible that making the critic Q(s, a) smooth around $\epsilon-neighbor$ of s can resist the attackers. In another word, with the small noise of the observation, if Q-function still recommends the same best action as the clean observation, it can resist the attacker. This smooth Q-function is then distilled to the policy when performing the \textit{policy improvement} resulting in a robust policy against attacks. This defense is referred to as the \textit{Critic defense} and the Q-function is learned by the below objective: 
\begin{equation}
   \min_Q J(Q_{\phi})+ \lambda_Q L(Q_{\phi}; A),
\end{equation}
where $J(Q_{\phi})$ is the conventional policy evaluation operator and $L(Q_{\phi}; A)$ is the augmented defense regularization with the weight $\lambda_Q$ to control the strength of regularization. The formula of $L(Q_{\phi}; A)$ is as followed:
\begin{equation}
L(Q_{\phi};  A) = \mathbb{E}_{(s,a) \sim D, \tilde{s} \sim A} \left[ (Q_{\phi}(\tilde{s},a) - Q_{\phi}(s,a))^2 \right],
\end{equation}      
where $A$ is the selected attack (e.g., critic attack or random attack ...) is used for generate $\tilde{s}$. For instance, the critic attack generate  $\tilde{s}$ as follow:
\begin{equation}
    \tilde{s} = \argmin_{\tilde{s} \in B(s,\epsilon)} Q_{\phi}(s, \pi_{\theta}(\tilde{s})).
\end{equation}
\textbf{Actor defense:} An actor that cannot output a very different action given just a small observation perturbation is a hard opponent of actor attacks. Following this philosophy, we  propose to smooth the actor so that a small perturbed observation can not produce a very different action compared to the clean one. This defense is referred to as the \textit{Actor Defense} and the actor is learned by the below objective:    
\begin{align}
    \min_\pi  J(\pi_{\theta})+ \lambda_{\pi} L(\pi_{\theta}; A),
\end{align}
where $J(\pi_{\theta})$ is the conventional policy improvement operator and $L(\pi_{\theta})$ is the defense regularization with the weight $\lambda_{\pi}$ to control the strength of regularization. The formula of  $L(\pi_{\theta}; A)$ is as following:
    \begin{equation}
    L(\pi_{\theta}; A) = \mathbb{E}_{s \sim D, \tilde{s} \sim A} \left[ (\pi_{\theta}(\tilde{s}) - \pi_{\theta}(s))^2 \right],  
    \end{equation}
where A is the selected training attack. One example is \textit{Actor Attack} which generates the perturbation as follow: 
\begin{equation}
    \tilde{s} = \argmax_{\tilde{s} \in B(s,\epsilon)} (\pi_{\theta}(\tilde{s}) - \pi_{\theta}(s))^2.  
\end{equation}

\section{Experiments}

\subsection{Experiment setup}
The proposed methods are evaluated on the D4RL \cite{fu2020d4rl} benchmark of OpenAI Gym MojoCo tasks, which consists of various datasets. We especially focus on three well-known tasks: Hopper-Hop, Half-Cheetah, and Walker-walk. For each task, three datasets will be evaluated: Expert, Medium-Replay, and Medium-Expert. These selections were made due to their significance in reinforcement learning research, offering a spectrum of complexities and dynamics. Such diversity enables gauging the generalization and robustness of proposed methods across varied settings.

 For implementation, TD3+BC, CQL, and BCQ are trained in accordance with their respective papers \cite{fujimoto2019off,kumar2020conservative,wu2019behavior}, with the exception of the following modifications. The critic network is an MLP consisting of two 256-d hidden layers followed by ReLU activations except for the last layer. The actor network has a similar architecture to the critic network. The only difference is that the activation in the last layer is Tanh. Unless specified, the default hyper-parameters of the algorithm are as follows: the actor and critic are trained using Adam optimizer \cite{kingma2014adam} with default parameters and a learning rate of 3e-4. A mini-batch size of 256 is used for training. The target networks and the actor updates are performed every two critic updates. Adversarial examples are generated by performing 5-steps PGD \cite{madry2017towards} with $\epsilon = 0.05$ and step size is $0.01$. The networks are trained with 500k iterations. For training the Q-function for Robust Critic Attack, the interaction budget with test environment is limited to 10000 transitions.  $\lambda_{\mu}$ and $\lambda_Q$ are searched in the range of $\{0.1, 0.5, 1, 5, 10\}$, and the best results are reported after normalized using d4rl scores, which provide a measure of performance relative to expert
and random scores. The normalization formula is as follows:
\begin{equation}
score_{normalized} = 100* \frac{score - score_{random} }{ score_{expert} - score_{random}}.  
\end{equation}

For evaluation, each method is trained using 5 different seeds and the aggregated performance is reported over trained models. We assess the algorithm using Interquartile Mean (IQM)\cite{agarwal2021deep}, a choice known for delivering high-confidence results even with a limited number of runs. In brief, IQM is the mean of the middle 50\% of runs in a consider set of runs, and is resistant to outliers. The IQM is one of the statistical tools proposed to increase the field’s confidence in reported results with a handful of runs. For reference, we also provide the Mean and Median metrics, which are well-established in the community. It is important to stress that, particularly in settings with a restricted number of runs, the Mean and Median metrics exhibit inconsistency, further accentuating IQM as the superior choice in such scenarios.

\begin{figure}[tbp]
% \vskip -1.5cm
\centering
\subfigure[Aggregated Performance of defenses in expert dataset.]{\includegraphics[width=0.9\linewidth]{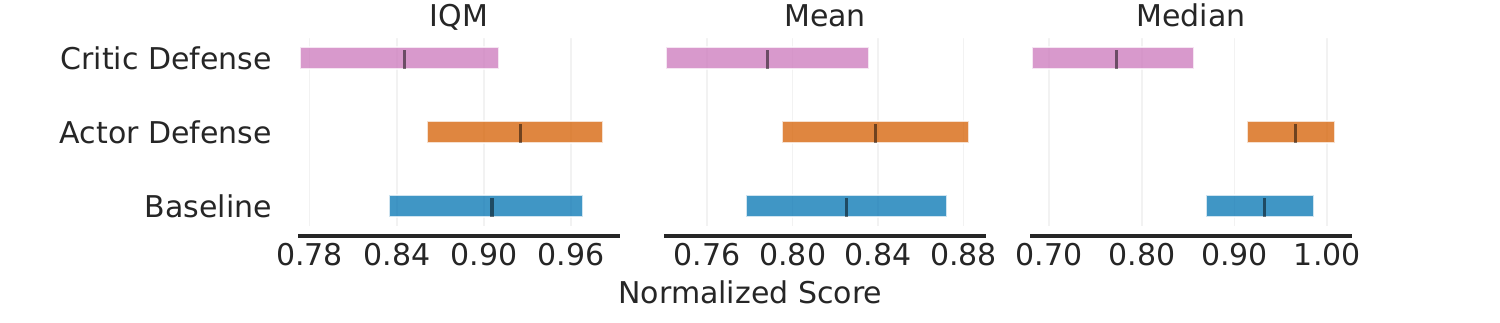}\label{fig:Defense_expert}}
\subfigure[Aggregated Performance of defenses in Medium-Expert dataset.]{\includegraphics[width=0.9\linewidth]{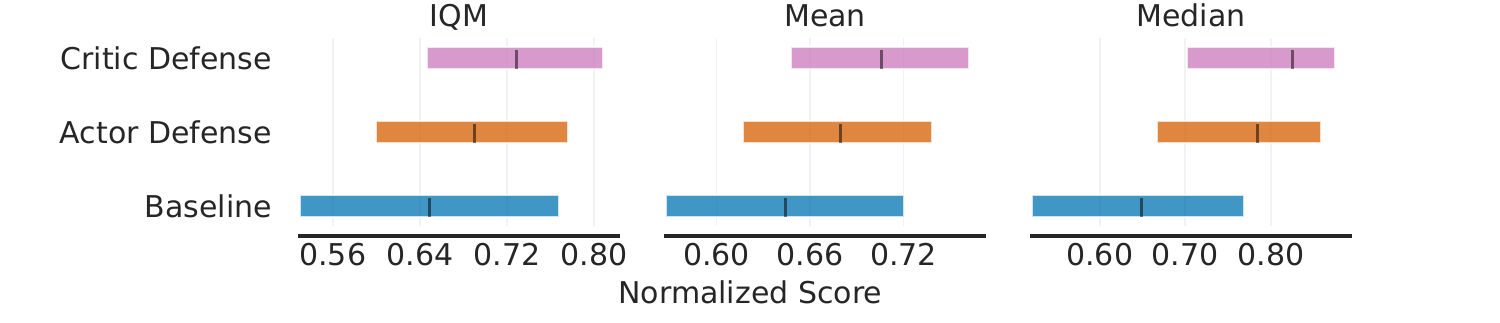}\label{fig:Defense_mediumexpert}}
\subfigure[Aggregated Performance of defenses in Medium-Replay dataset.]{\includegraphics[width=0.9\linewidth]{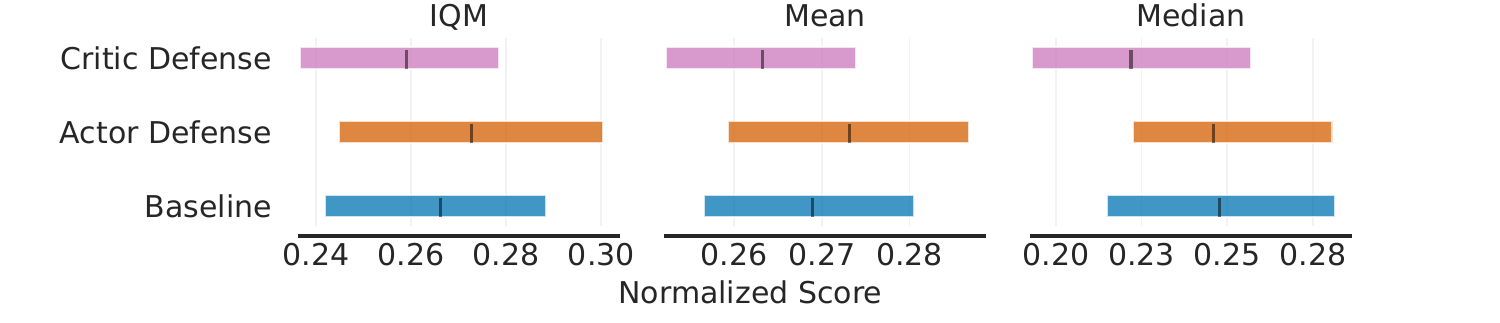}\label{fig:Defense_mediumreplay}}
\subfigure[Overall performance of defenses among three datasets.]{\includegraphics[width=0.9\linewidth]{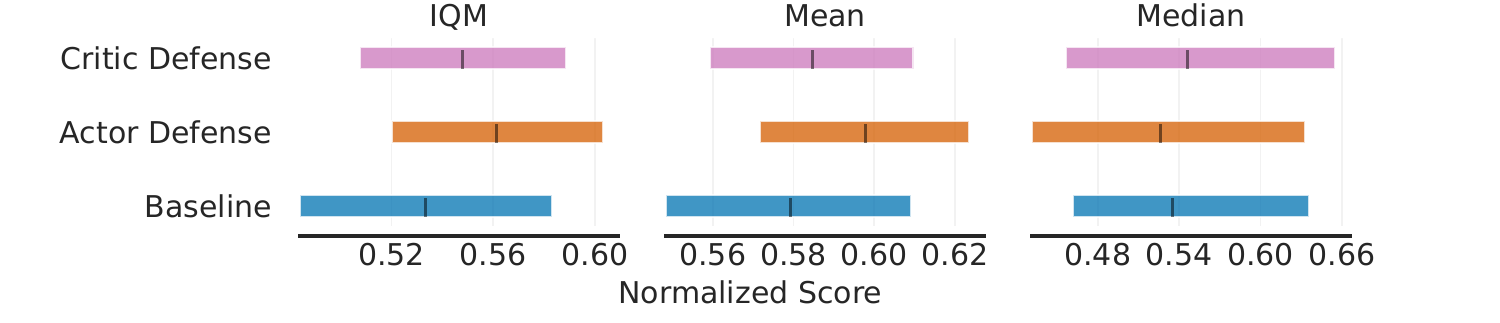}\label{fig:Defense_Overal}}
\caption{Aggregated Performance of Defenses with 95\% Confidence Intervals (CIs). The aggregated performances are calculated as the average performance over tasks and attacks, specifically on certain types of datasets. The overall performance is further averaged over different datasets.}
\label{fig:DefensePerformance}
% \vskip -0.7cm
\end{figure}

\subsection{Experimental Results}

\begin{table*}[tbp]
\centering
{
\begin{tabular}{@{}|l|c|c|c|c|c|@{}}
\toprule
& \multicolumn{1}{l|}{Clean} & \multicolumn{1}{l|}{Random Attack} & \multicolumn{1}{l|}{Critic Attack} & \multicolumn{1}{l|}{Actor Attack} & \multicolumn{1}{l|}{Robust Critic Attack} \\ \midrule
TD3+BC & 73.46                      & 72.98 (\textcolor{red}{-1\%})                       & 43.69 (\textcolor{red}{-41\%})                      & 70.10 (\textcolor{red}{-5\%})                      & 48.15 (\textcolor{red}{-34\%})                            \\ \midrule
CQL    & 55.84                      & 54.92 (\textcolor{red}{-2\%})                       & 36.66 (\textcolor{red}{-34\%})                      & 42.89 (\textcolor{red}{-23\%})                     & 13.79 (\textcolor{red}{-75\%})                            \\ \midrule
BCQ    & 49.82                      & 48.67 (\textcolor{red}{-2\%})                         & 17.17 (\textcolor{red}{-66\%})                      & 41.77 (\textcolor{red}{-16\%})                     & 16.13 (\textcolor{red}{-68\%})                             \\ \bottomrule
\end{tabular}
}%
% \caption{Adversarial attacks IQM-aggregated performance on offline RL in the Walker Walk task using different datasets (expert, medium-expert, and medium-replay)}
\caption{The performance of adversarial attacks in multiple offline RL algorithms in Walker-Walk with three datasets including expert, medium-expert, and medium-replay. We report the IQM-aggregated performance for all three datasets.}
\label{tab:IQMAttackOfflineMethods}
\end{table*}

 We present the results including (1) comparison among attacks, (2) comparison among defenses, and (3) impact on the training cost.

% \begin{figure}[htp]
% \subfigure[Expert dataset performance.]{\includegraphics[width=0.8\linewidth]{figs/Attack_Expert.pdf}
% \label{fig:attack_expert}
% }
  
%  \subfigure[Medium dataset performance.]{ 
% \includegraphics[width=0.8\linewidth]{figs/Attack_MediumExpert.pdf}
%   \label{fig:attack_mediumexpert}
% }
% \subfigure[Medium-Replay dataset performance.] {   \includegraphics[width=0.8\linewidth]{figs/Attack_MediumReplay.pdf}
%   \label{fig:attack_mediumreplay}
%   }
% \subfigure[Overall Performance among three datasets.]
% {
%   \includegraphics[width=0.8\linewidth]{figs/Attack_Overal.pdf}
%   \label{fig:attack_overal}
% }
% \caption{Aggregated performance of Attackers with 95\% CIs, showcasing metrics IQM, MEAN, and MEDIUM. IQM, proven optimal in our experiment setting with a limited number of runs, provides key insights. Performances are aggregated across specific dataset types, averaging over tasks, and overall performance further averages across datasets.}
% \label{fig:attackPerformance}
% \vspace{-2cm}
% \end{figure}
\textbf{(1) Comparison among attacks:} 
\textbf{Attacks consistently degrade performance across various offline RL methods and datasets}. More specific, we train BCQ \cite{wu2019behavior}, CQL \cite{kumar2020conservative}, and TD3+BC \cite{fujimoto2021minimalist} using their standard training procedure in Walker-Walk with three corresponding datasets: Expert, Medium-Replay, and Medium-Expert. In the testing phase, we evaluate the trained model to various Attacks and report the IQM score. The scores are aggregated over three datasets, with five runs for each, and are presented in Table \ref{tab:IQMAttackOfflineMethods}. According to the result, it is evident that Random Attack is the least effective, while attacks based on critic or actor result in significant performance degradation up to -75\% for all considered offline RL methods. This degradation can be anticipated as deep neural networks are susceptible to input perturbations.

\textbf{Attacks degrade performance across various  tasks and datasets.} To be more specific, we further choose TD3-BC as baseline and evaluated under the attacks across tasks: Hopper-Hop, Half-Cheetah, and Walker-walk. For each task, three datasets will be evaluated: Expert, Medium-Replay, and Medium-Expert. IQM score are aggregated over three datasets and three tasks, with five runs for each. The IQM result demonstrates that without any attack, the baseline agent achieves a normalized score of 0.87. The Random Attack reduces the performance by -18.4\% (0.71/0.87), while the smarter attacks result in greater reduction: -39\% (0.53/0.87) for Actor Attack, -57\% (0.37/0.87) for Critic Attack, and -63\% (0.32/0.87) for Robust Critic Attack.  According to the result the ranking order of attacks (1) Robust Critic Attack, (2) Critic Attack, (3) Actor Attack, and (4) Random Attack. These results can be elucidated by considering the nature of the attacks. The Random Attack exhibited lesser efficacy due to its indiscriminate nature. Conversely, the Actor and Critic attacks strategically targeted the core components of RL algorithms, thereby enhancing their effectiveness. Specifically, the Critic Attack aimed at the long-term return, reflecting a more farsighted approach compared to the Actor Attack, resulting in a slight improvement. Finally, the Robust Critic Attack, leveraging an accurate Q function, emerged as the most effective, surpassing even the Critic Attack in its efficacy.
% The order of attacks is clearly separable under confidence intervals (CIs), indicating high reliability. For full results, please refer to the Appendix.

\textbf{(2) Comparison among defenses:} Defenses are applied to enhance the robustness of TD3-BC, called Robust TD3-BC, on various tasks (Hopper-Hop, Half-Cheetah, and Walker-walk) and datasets (Expert, Medium-Replay, and Medium-Expert) in our proposed framework for robust offline RL training. The Robust TD3-BC is then evaluated with various attacks in the test environment and IQM is reported. There are IQM aggregated performances on certain types of datasets where the performances are averaged out over tasks and the overall performance where the performances are further averaged out over datasets. The aggregated performance is shown in Figure \ref{fig:DefensePerformance}, where Figure \ref{fig:Defense_expert}, Figure \ref{fig:Defense_mediumexpert}, Figure \ref{fig:Defense_mediumreplay} display dataset-specific performance and Figure \ref{fig:Defense_Overal} displays the average performance of defenses among three datasets.

Overall, the Actor Defense is the most effective under attacks, with a performance improvement of +11\% (0.57/0.53) over the baseline, while the Critic Defense shows a slightly lower improvement of +10\% (0.55/0.53). The performance gap between the best defense and the clean environment is significant (0.57-0.87), indicating room for further improvement in defenses.

When examining each dataset in detail, the Actor Defense consistently outperforms the baseline for all datasets, which aligns with the overall performance. Notably, the Critic Defense does not improve the performance of the baseline on the Expert dataset and Medium-Replay dataset, as suggested by the overall performance. However, on the Medium-Expert dataset, the Critic Defense does show performance improvement. This suggests that the proposed Critic Defense may not be effective with low-return data (Medium-Replay dataset) or high-return but narrow and biased data (Expert dataset), but performs well with wide distribution and high-return data (Medium-Expert).

\textbf{(3) Impact on the training cost:} While the proposed method seems to improve the model's robustness, questions may arise regarding its potential impact on the training costs. To clarify, there is an associated increase in training costs due to the generation of perturbations during training—a common aspect in the development of robust models. For reference, we provide the wall-clock training times for 0.5 million iterations on TD3-BC as follows: TD3-BC: 3.05 hours, TD3-BC-actor defense: 4.55 hours, TD3-BC-critic defense: 5.56 hours. Minimizing the training cost associated with robust training could be a potential avenue for future research.

% \textbf{Impact on the training cost} While the proposed method seems to improve the model's robustness, questions may arise regarding its potential impact on the number of parameters and training costs. To clarify, our defense strategy relies solely on incorporating additional regularization into the training of the policy/Q-value function, thereby introducing no extra parameters. However, there is an associated increase in training costs due to the generation of perturbations during training—a common aspect in the development of robust models. For reference, we provide the wall-clock training times for 0.5 million iterations on TD3-BC as follows: TD3-BC: 3.05 hours, TD3-BC-actor defense: 4.55 hours, TD3-BC-critic defense: 5.56 hours. Minimizing the training cost associated with robust training could be a potential avenue for future research.

\section{Conclusion}
In conclusion, this paper addresses the challenge of improving the robustness of offline RL models by leveraging advanced adversarial attacks and defenses. The proposed framework introduces attacks on the actor and critic components during training and proposes regularization-based defenses to enhance their robustness. Through extensive experiments on the D4RL benchmark, the framework is shown to be effective in enhancing the reliability of offline RL models. The results highlight the vulnerability of both actor and critic components to adversarial attacks and the effectiveness of the proposed defenses in mitigating their impact. Further research in this area has the potential to develop more robust and reliable RL models for real-world applications.

\section{Acknowledgement}
This work was supported by the Institute of Information \& communications Technology Planning \& Evaluation (IITP) grant funded by the Korean government(MSIT) (No.2022-0-00184, Development
and Study of AI Technologies to Inexpensively Conform to Evolving Policy on Ethics and No. 2021-0-01381, Development of Causal AI through Video Understanding and Reinforcement Learning, and Its Applications to Real Environments).

\bibliographystyle{splncs04}
\bibliography{refs}

\begin{thebibliography}{10}
\providecommand{\url}[1]{\texttt{#1}}
\providecommand{\urlprefix}{URL }
\providecommand{\doi}[1]{https://doi.org/#1}

\bibitem{agarwal2020optimistic}
Agarwal, R., Schuurmans, D., Norouzi, M.: An optimistic perspective on offline
  reinforcement learning. In: International Conference on Machine Learning. pp.
  104--114. PMLR (2020)

\bibitem{agarwal2021deep}
Agarwal, R., Schwarzer, M., Castro, P.S., Courville, A.C., Bellemare, M.: Deep
  reinforcement learning at the edge of the statistical precipice. Advances in
  neural information processing systems  \textbf{34},  29304--29320 (2021)

\bibitem{an2021uncertainty}
An, G., Moon, S., Kim, J.H., Song, H.O.: Uncertainty-based offline
  reinforcement learning with diversified q-ensemble. Advances in neural
  information processing systems  \textbf{34},  7436--7447 (2021)

\bibitem{bai2022pessimistic}
Bai, C., Wang, L., Yang, Z., Deng, Z., Garg, A., Liu, P., Wang, Z.: Pessimistic
  bootstrapping for uncertainty-driven offline reinforcement learning. ICLR
  (2022)

\bibitem{brandfonbrener2022incorporating}
Brandfonbrener, D., Combes, R.T.d., Laroche, R.: Incorporating explicit
  uncertainty estimates into deep offline reinforcement learning. arXiv
  preprint arXiv:2206.01085  (2022)

\bibitem{brandfonbrener2021offline}
Brandfonbrener, D., Whitney, W., Ranganath, R., Bruna, J.: Offline rl without
  off-policy evaluation. Advances in Neural Information Processing Systems
  \textbf{34},  4933--4946 (2021)

\bibitem{chen2019information}
Chen, J., Jiang, N.: Information-theoretic considerations in batch
  reinforcement learning. In: International Conference on Machine Learning. pp.
  1042--1051. PMLR (2019)

\bibitem{chen2020bail}
Chen, X., Zhou, Z., Wang, Z., Wang, C., Wu, Y., Ross, K.: Bail: Best-action
  imitation learning for batch deep reinforcement learning. Advances in Neural
  Information Processing Systems  \textbf{33},  18353--18363 (2020)

\bibitem{devlin2018bert}
Devlin, J., Chang, M.W., Lee, K., Toutanova, K.: Bert: Pre-training of deep
  bidirectional transformers for language understanding. arXiv preprint
  arXiv:1810.04805  (2018)

\bibitem{fu2020d4rl}
Fu, J., Kumar, A., Nachum, O., Tucker, G., Levine, S.: D4rl: Datasets for deep
  data-driven reinforcement learning. arXiv preprint arXiv:2004.07219  (2020)

\bibitem{fujimoto2021minimalist}
Fujimoto, S., Gu, S.S.: A minimalist approach to offline reinforcement
  learning. Advances in neural information processing systems  \textbf{34},
  20132--20145 (2021)

\bibitem{fujimoto2019off}
Fujimoto, S., Meger, D., Precup, D.: Off-policy deep reinforcement learning
  without exploration. In: International Conference on Machine Learning. pp.
  2052--2062 (2019)

\bibitem{fujimoto2018addressing}
Fujimoto, S., Van~Hoof, H., Meger, D.: Addressing function approximation error
  in actor-critic methods. Proc. Int. Conf. Mach. Learn (ICML) pp. 1582--1591
  (2018)

\bibitem{he2016deep}
He, K., Zhang, X., Ren, S., Sun, J.: Deep residual learning for image
  recognition. In: Proceedings of the IEEE conference on computer vision and
  pattern recognition. pp. 770--778 (2016)

\bibitem{huang2017adversarial}
Huang, S., Papernot, N., Goodfellow, I., Duan, Y., Abbeel, P.: Adversarial
  attacks on neural network policies. arXiv preprint arXiv:1702.02284  (2017)

\bibitem{jin2021pessimism}
Jin, Y., Yang, Z., Wang, Z.: Is pessimism provably efficient for offline rl?
  In: International Conference on Machine Learning. pp. 5084--5096. PMLR (2021)

\bibitem{kingma2014adam}
Kingma, D.P., Ba, J.: Adam: A method for stochastic optimization. arXiv
  preprint arXiv:1412.6980  (2014)

\bibitem{kos2017delving}
Kos, J., Song, D.: Delving into adversarial attacks on deep policies. arXiv
  preprint arXiv:1705.06452  (2017)

\bibitem{kostrikov2022offline}
Kostrikov, I., Nair, A., Levine, S.: Offline reinforcement learning with
  implicit q-learning. ICLR  (2022)

\bibitem{krizhevsky2017imagenet}
Krizhevsky, A., Sutskever, I., Hinton, G.E.: Imagenet classification with deep
  convolutional neural networks. Communications of the ACM  \textbf{60}(6),
  84--90 (2017)

\bibitem{kumar2019stabilizing}
Kumar, A., Fu, J., Soh, M., Tucker, G., Levine, S.: Stabilizing off-policy
  q-learning via bootstrapping error reduction. Advances in Neural Information
  Processing Systems  \textbf{32} (2019)

\bibitem{kumar2020conservative}
Kumar, A., Zhou, A., Tucker, G., Levine, S.: Conservative q-learning for
  offline reinforcement learning. Advances in Neural Information Processing
  Systems  \textbf{33},  1179--1191 (2020)

\bibitem{levine2016end}
Levine, S., Finn, C., Darrell, T., Abbeel, P.: End-to-end training of deep
  visuomotor policies. The Journal of Machine Learning Research
  \textbf{17}(1),  1334--1373 (2016)

\bibitem{levine2020offline}
Levine, S., Kumar, A., Tucker, G., Fu, J.: Offline reinforcement learning:
  Tutorial, review, and perspectives on open problems. arXiv preprint
  arXiv:2005.01643  (2020)

\bibitem{lin2017tactics}
Lin, Y.C., Hong, Z.W., Liao, Y.H., Shih, M.L., Liu, M.Y., Sun, M.: Tactics of
  adversarial attack on deep reinforcement learning agents. arXiv preprint
  arXiv:1703.06748  (2017)

\bibitem{luu2022utilizing}
Luu, T.M., Nguyen, T., Vu, T., Yoo, C.D.: Utilizing skipped frames in action
  repeats for improving sample efficiency in reinforcement learning. IEEE
  Access  \textbf{10},  64965--64975 (2022)

\bibitem{luu2022visual}
Luu, T.M., Vu, T., Nguyen, T., Yoo, C.D.: Visual pretraining via contrastive
  predictive model for pixel-based reinforcement learning. Sensors
  \textbf{22}(17), ~6504 (2022)

\bibitem{madry2017towards}
Madry, A., Makelov, A., Schmidt, L., Tsipras, D., Vladu, A.: Towards deep
  learning models resistant to adversarial attacks. arXiv preprint
  arXiv:1706.06083  (2017)

\bibitem{mnih2015human}
Mnih, V., Kavukcuoglu, K., Silver, D., Rusu, A.A., Veness, J., Bellemare, M.G.,
  Graves, A., Riedmiller, M., Fidjeland, A.K., Ostrovski, G., et~al.:
  Human-level control through deep reinforcement learning. Nature
  \textbf{518}(7540),  529--533 (2015)

\bibitem{munos2016safe}
Munos, R., Stepleton, T., Harutyunyan, A., Bellemare, M.: Safe and efficient
  off-policy reinforcement learning. Advances in neural information processing
  systems  \textbf{29} (2016)

\bibitem{nguyen2021robust}
Nguyen, T., Luu, T., Pham, T., Rakhimkul, S., Yoo, C.D.: Robust maml:
  Prioritization task buffer with adaptive learning process for model-agnostic
  meta-learning. In: ICASSP 2021-2021 IEEE International Conference on
  Acoustics, Speech and Signal Processing (ICASSP). pp. 3460--3464. IEEE (2021)

\bibitem{nguyen2023fast}
Nguyen, T., Luu, T., Yoo, C.D.: Fast and memory-efficient uncertainty-aware
  framework for offline reinforcement learning with rank one mimo q network.
  In: IROS 2023 Workshop on Policy Learning in Geometric Spaces. IROS 2023
  Workshop (2023)

\bibitem{nguyen2021sample}
Nguyen, T., Luu, T.M., Vu, T., Yoo, C.D.: Sample-efficient reinforcement
  learning representation learning with curiosity contrastive forward dynamics
  model. In: 2021 IEEE/RSJ International Conference on Intelligent Robots and
  Systems (IROS). pp. 3471--3477. IEEE (2021)

\bibitem{nguyen2023dimcl}
Nguyen, T., Pham, T.X., Zhang, C., Luu, T.M., Vu, T., Yoo, C.D.: Dimcl:
  Dimensional contrastive learning for improving self-supervised learning. IEEE
  Access  \textbf{11},  21534--21545 (2023)

\bibitem{pattanaik2018robust}
Pattanaik, A., Tang, Z., Liu, S., Bommannan, G., Chowdhary, G.: Robust deep
  reinforcement learning with adversarial attacks. AAMAS  (2018)

\bibitem{rashidinejad2021bridging}
Rashidinejad, P., Zhu, B., Ma, C., Jiao, J., Russell, S.: Bridging offline
  reinforcement learning and imitation learning: A tale of pessimism. Advances
  in Neural Information Processing Systems  \textbf{34},  11702--11716 (2021)

\bibitem{schulman2015trust}
Schulman, J., Levine, S., Abbeel, P., Jordan, M., Moritz, P.: Trust region
  policy optimization. In: Proc. Int. Conf. Mach. Learn (ICML). pp. 1889--1897
  (2015)

\bibitem{shen2020deep}
Shen, Q., Li, Y., Jiang, H., Wang, Z., Zhao, T.: Deep reinforcement learning
  with robust and smooth policy. In: ICML. pp. 8707--8718 (2020)

\bibitem{silver2016mastering}
Silver, D., Huang, A., Maddison, C.J., Guez, A., Sifre, L., Van Den~Driessche,
  G., Schrittwieser, J., Antonoglou, I., Panneershelvam, V., Lanctot, M.,
  et~al.: Mastering the game of go with deep neural networks and tree search.
  Nature  \textbf{529}(7587),  484–489 (2016)

\bibitem{sun2021strongest}
Sun, Y., Zheng, R., Liang, Y., Huang, F.: Who is the strongest enemy? towards
  optimal and efficient evasion attacks in deep rl. ICLR  (2022)

\bibitem{torabi2018behavioral}
Torabi, F., Warnell, G., Stone, P.: Behavioral cloning from observation. arXiv
  preprint arXiv:1805.01954  (2018)

\bibitem{vaswani2017attention}
Vaswani, A., Shazeer, N., Parmar, N., Uszkoreit, J., Jones, L., Gomez, A.N.,
  Kaiser, {\L}., Polosukhin, I.: Attention is all you need. Advances in neural
  information processing systems  \textbf{30} (2017)

\bibitem{vu2022softgroup}
Vu, T., Kim, K., Luu, T.M., Nguyen, T., Yoo, C.D.: Softgroup for 3d instance
  segmentation on point clouds. In: Proceedings of the IEEE/CVF Conference on
  Computer Vision and Pattern Recognition. pp. 2708--2717 (2022)

\bibitem{vu2023scalable}
Vu, T., Kim, K., Nguyen, T., Luu, T.M., Kim, J., Yoo, C.D.: Scalable softgroup
  for 3d instance segmentation on point clouds. IEEE Transactions on Pattern
  Analysis and Machine Intelligence  (2023)

\bibitem{wang2020critic}
Wang, Z., Novikov, A., Zolna, K., Merel, J.S., Springenberg, J.T., Reed, S.E.,
  Shahriari, B., Siegel, N., Gulcehre, C., Heess, N., et~al.: Critic
  regularized regression. Advances in Neural Information Processing Systems
  \textbf{33},  7768--7778 (2020)

\bibitem{wu2019behavior}
Wu, Y., Tucker, G., Nachum, O.: Behavior regularized offline reinforcement
  learning. arXiv preprint arXiv:1911.11361  (2019)

\bibitem{xie2021bellman}
Xie, T., Cheng, C.A., Jiang, N., Mineiro, P., Agarwal, A.: Bellman-consistent
  pessimism for offline reinforcement learning. Advances in neural information
  processing systems  \textbf{34},  6683--6694 (2021)

\bibitem{yang2022rorl}
Yang, R., Bai, C., Ma, X., Wang, Z., Zhang, C., Han, L.: Rorl: Robust offline
  reinforcement learning via conservative smoothing. Advances in Neural
  Information Processing Systems  \textbf{35},  23851--23866 (2022)

\bibitem{yin2022near}
Yin, M., Duan, Y., Wang, M., Wang, Y.X.: Near-optimal offline reinforcement
  learning with linear representation: Leveraging variance information with
  pessimism. arXiv preprint arXiv:2203.05804  (2022)

\bibitem{zhang2021robust}
Zhang, H., Chen, H., Boning, D., Hsieh, C.J.: Robust reinforcement learning on
  state observations with learned optimal adversary. ICLR  (2021)

\bibitem{zhang2020robust}
Zhang, H., Chen, H., Xiao, C., Li, B., Liu, M., Boning, D., Hsieh, C.J.: Robust
  deep reinforcement learning against adversarial perturbations on state
  observations. Advances in Neural Information Processing Systems  \textbf{33},
   21024--21037 (2020)

\bibitem{zhang2022corruption}
Zhang, X., Chen, Y., Zhu, X., Sun, W.: Corruption-robust offline reinforcement
  learning. In: International Conference on Artificial Intelligence and
  Statistics. pp. 5757--5773. PMLR (2022)

\end{thebibliography}

\end{document}

\section{Appendix}
\subsection{IQM performance of TD3-BC (without robust training) under various attacks}

\begin{figure}[H]
\scriptsize
\centering
\subfigure[Expert dataset performance.]{\includegraphics[width=0.8\linewidth]{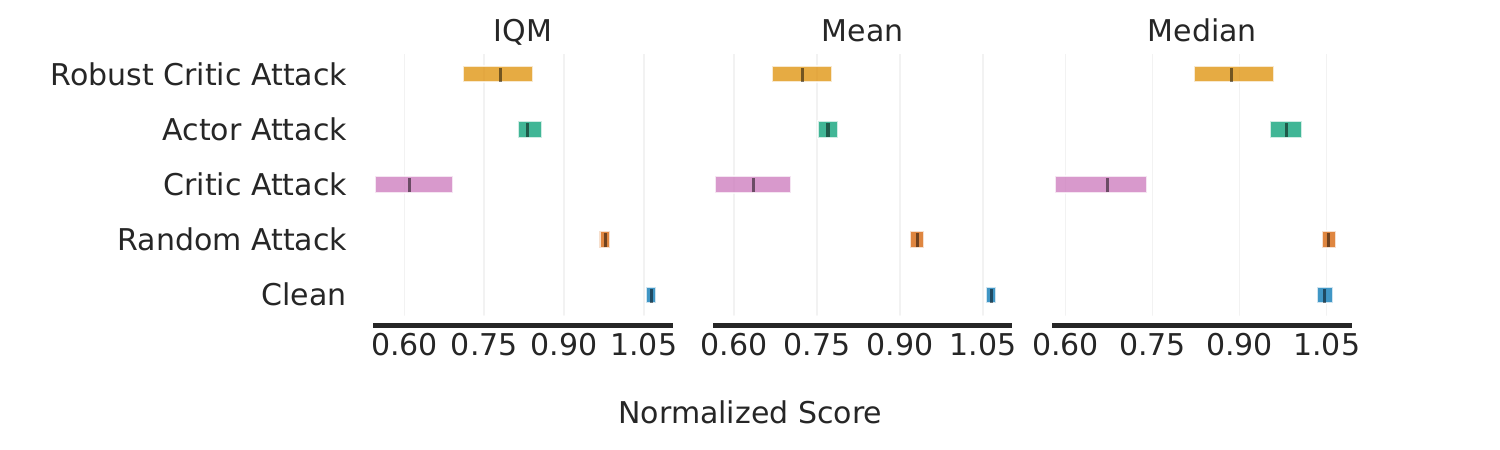}
\label{fig:attack_expert}
}
 \subfigure[Medium dataset performance.]{ 
\includegraphics[width=0.8\linewidth]{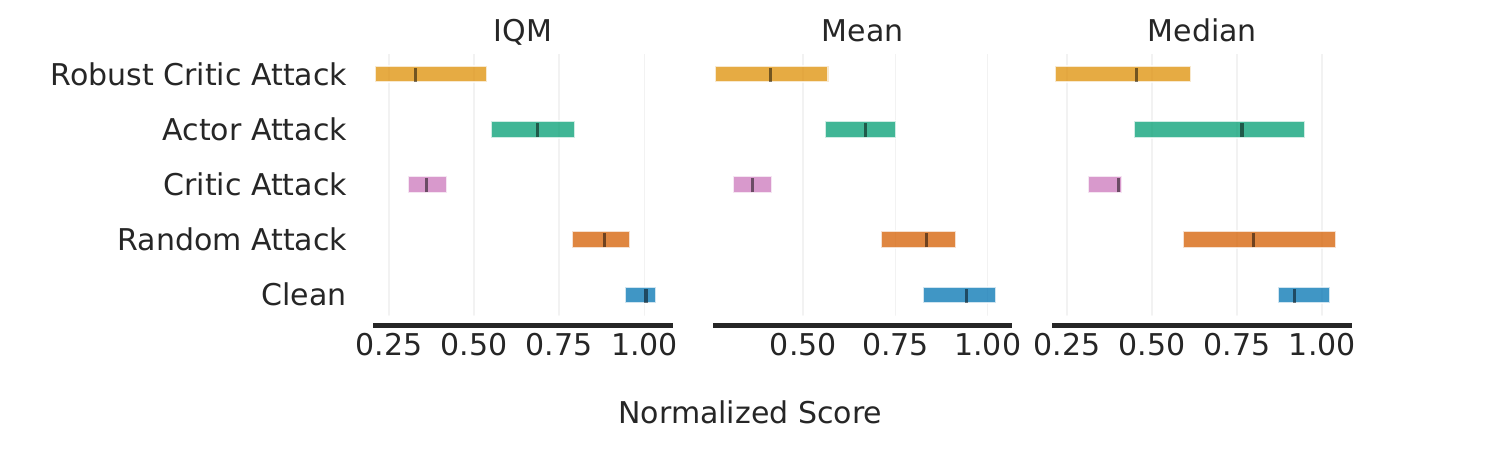}
  \label{fig:attack_mediumexpert}
}
\subfigure[Medium-Replay dataset performance.] {   \includegraphics[width=0.8\linewidth]{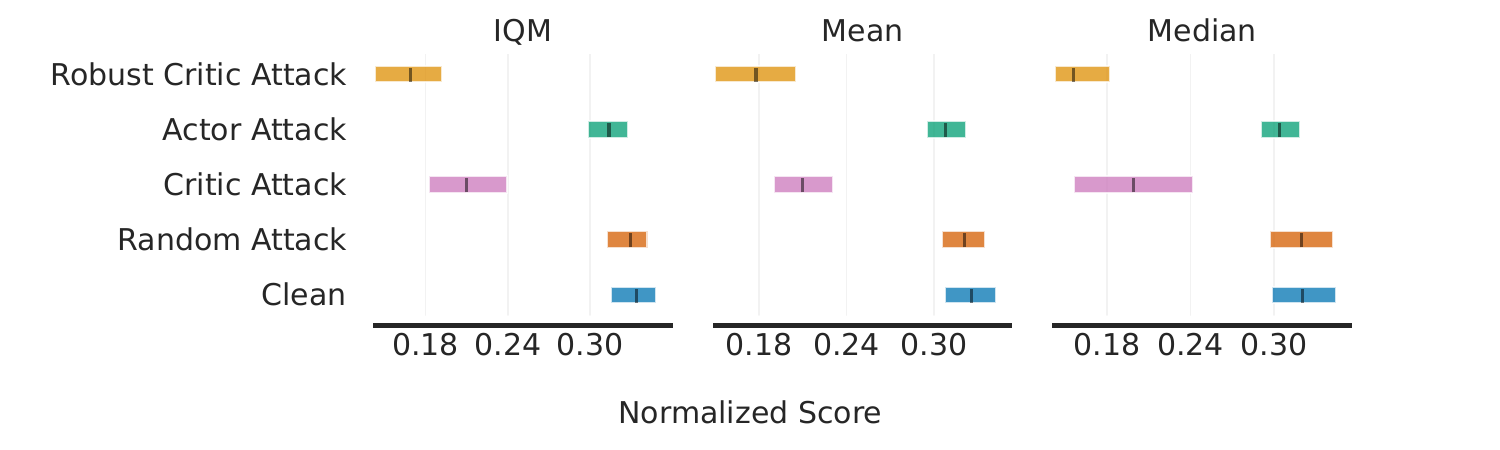}
  \label{fig:attack_mediumreplay}
  }
\subfigure[Overall Performance among three datasets.]
{
  \includegraphics[width=0.8\linewidth]{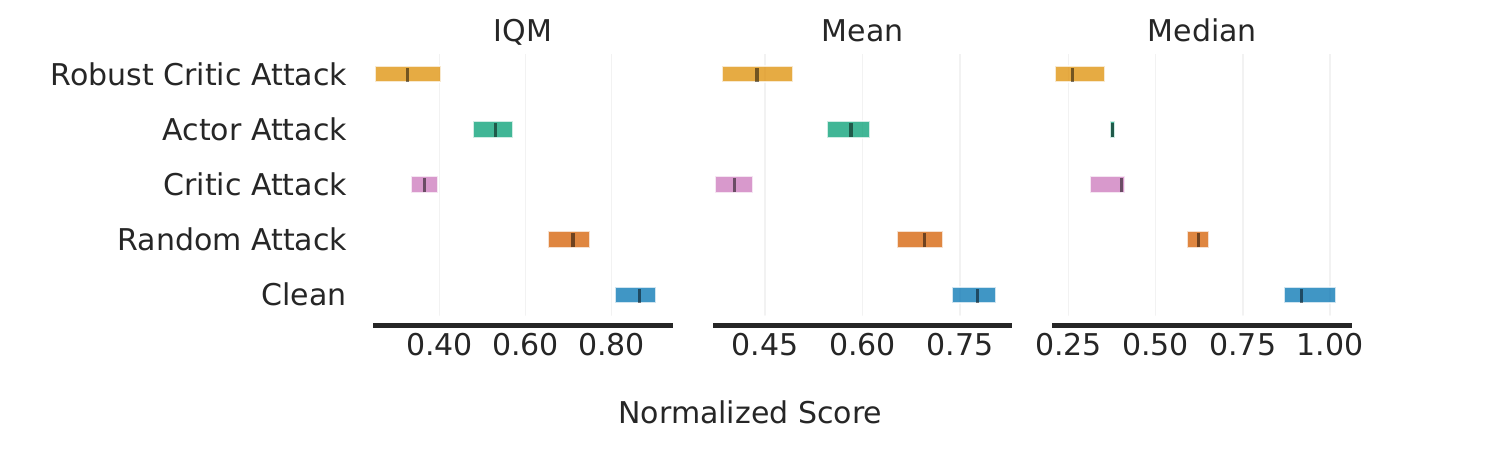}
  \label{fig:attack_overal}
}
\caption{Aggregated performance of Attackers with 95\% CIs, showcasing metrics IQM, MEAN, and MEDIUM. IQM, proven optimal in our experiment setting with a limited number of runs, provides key insights. Performances are aggregated across specific dataset types, averaging over tasks, and overall performance further averages across datasets.}
\label{fig:attackPerformance}
\end{figure}

\subsection{Full experiment score of TD3-BC baseline}

\begin{table*}[]
\tiny
\centering
\begin{tabular}{@{}|l|l|c|c|c|c|c|@{}}
\toprule
Task                          & Method                & Clean               & Random Attack       & Critic Attack       & Actor Attack        & Robust Critic Attack \\ \midrule
\multirow{3}{*}{Walker walk}  & TD3BC                 & 22.4+5.35           & 21.61+4.31          & 19.89+5.06          & 24.3+4.2            & 15.6+2.5             \\ \cmidrule(l){2-7} 
                              & TD3BC+ Critic Defense & \textit{23.13+4.52} & \textit{23.67+4.17} & \textit{15.89+0.82} & \textit{21.46+2.84} & 13.63+2.01           \\ \cmidrule(l){2-7} 
                              & TD3BC+ Actor Defense  & \textit{26.93+7.16} & \textit{29.17+7.85} & \textit{17.27+3.78} & \textit{29.59+8.25} & 13.66+2.98           \\ \midrule
\multirow{3}{*}{Hopper hop}   & TD3BC                 & \textit{32.03+2.72} & \textit{31.97+2.63} & \textit{13.85+4.23} & \textit{30.42+1.62} & 15.59+3.23           \\ \cmidrule(l){2-7} 
                              & TD3BC+ Critic Defense & \textit{30.53+1.73} & \textit{29.61+2.35} & \textit{10.43+2.48} & \textit{28.09+1.61} & 12.37+4              \\ \cmidrule(l){2-7} 
                              & TD3BC+ Actor Defense  & \textit{32.44+2.74} & \textit{31.6+2.71}  & \textit{12.66+1.33} & \textit{29.76+1.7}  & 16.47+2.59           \\ \midrule
\multirow{3}{*}{Half Cheetah} & TD3BC                 & 43.17+0.37          & 42.62+0.41          & 29.23+2             & 37.64+0.74          & 22.18+8.65           \\ \cmidrule(l){2-7} 
                              & TD3BC+ Critic Defense & 42.53+0.99          & 42.06+0.73          & 29.99+3.24          & 37.74+1.11          & 28.69+5.83           \\ \cmidrule(l){2-7} 
                              & TD3BC+ Actor Defense  & 43.23+0.31          & 42.75+0.31          & 26.88+3.07          & 38.31+0.88          & 21.35+7.48           \\ \midrule
\multirow{3}{*}{MEAN}         & TD3BC                 & 32.53+2.81          & 32.07+2.45          & 20.99+3.76          & 30.79+2.19          & 17.79+4.79           \\ \cmidrule(l){2-7} 
                              & TD3BC+ Critic Defense & 32.06+2.41          & 31.78+2.42          & 18.77+2.18          & 29.09+1.85          & 18.23+3.95           \\ \cmidrule(l){2-7} 
                              & TD3BC+ Actor Defense  & 34.2+3.4            & 34.51+3.62          & 18.94+2.73          & 32.55+3.61          & 17.16+4.35           \\ \bottomrule
\end{tabular}
\caption{The effectiveness of adversarial attacks the baseline TD3-BC and TD3-BC with defenses on Medium-Replay dataset. Following previous work, Mean and Standard deviation are reported. Each experiment are performed on 5 different seeds. }
\label{tab:mediumexpert_performance}
\end{table*}

\begin{table*}[]
\tiny
\centering
\begin{tabular}{@{}|l|l|c|c|c|c|c|@{}}
\toprule
Task                          & Method                & Clean                & Random Attack        & Critic Attack        & Actor Attack         & Robust Critic Attack \\ \midrule
\multirow{3}{*}{Walker walk}  & TD3BC                 & 79.14+35.07          & 79.79+37             & 43.86+15.33          & 76.4+32.41           & 45.41+28.13          \\ \cmidrule(l){2-7} 
                              & TD3BC+ Critic Defense & \textit{95.62+6.7}   & \textit{95.76+6.53}  & \textit{60.22+8.18}  & \textit{90.95+4.87}  & 70.24+17.53          \\ \cmidrule(l){2-7} 
                              & TD3BC+ Actor Defense  & \textit{88.09+17.52} & \textit{89.5+13.53}  & \textit{70.34+9.26}  & \textit{89.48+11.67} & 54.67+37.65          \\ \midrule
\multirow{3}{*}{Hopper hop}   & TD3BC                 & \textit{111.8+0.88}  & \textit{110.26+1.52} & \textit{24.91+9.61}  & \textit{92.37+8.38}  & 54.19+44.51          \\ \cmidrule(l){2-7} 
                              & TD3BC+ Critic Defense & \textit{112.12+0.2}  & \textit{111.77+0.67} & \textit{45.26+26.45} & \textit{89.08+6.46}  & 62.71+39.24          \\ \cmidrule(l){2-7} 
                              & TD3BC+ Actor Defense  & \textit{112.15+0.17} & \textit{111.84+0.7}  & \textit{37.79+24.8}  & \textit{103.14+6.39} & 58.51+33.43          \\ \midrule
\multirow{3}{*}{Half Cheetah} & TD3BC                 & 91.94+6.17           & 60.1+2.79            & 40.16+1.32           & 32.15+1.85           & 23.82+7.88           \\ \cmidrule(l){2-7} 
                              & TD3BC+ Critic Defense & 80.69+8.7            & 48.48+5.8            & 42.62+2.22           & 31.43+2.6            & 22.27+7.7            \\ \cmidrule(l){2-7} 
                              & TD3BC+ Actor Defense  & 66.9+7.37            & 44.28+3.99           & 38.5+1.88            & 32.02+0.93           & 22.57+7.36           \\ \midrule
\multirow{3}{*}{MEAN}         & TD3BC                 & 94.29+14.04          & 83.38+13.77          & 36.31+8.75           & 66.98+14.22          & 41.14+26.84          \\ \cmidrule(l){2-7} 
                              & TD3BC+ Critic Defense & 96.14+5.2            & 85.34+4.34           & 49.37+12.28          & 70.48+4.64           & 51.74+21.49          \\ \cmidrule(l){2-7} 
                              & TD3BC+ Actor Defense  & 89.04+8.35           & 81.88+6.07           & 48.87+11.98          & 74.88+6.33           & 45.25+26.15          \\ \bottomrule
\end{tabular}

\caption{The effectiveness of adversarial attacks the baseline TD3-BC and TD3-BC with defenses on Medium-Expert dataset.}
\label{tab:mediumReplay_performance}
\end{table*}

\begin{table*}[]
\tiny
\centering
\begin{tabular}{@{}|l|l|c|c|c|c|c|@{}}
\toprule
Task                          & Method                & Clean                & Random Attack        & Critic Attack        & Actor Attack         & Robust Critic Attack \\ \midrule
\multirow{3}{*}{Walker walk}  & TD3BC                 & 104.69+1.61          & 105.38+1.42          & 67.27+8.77           & 100.29+4.65          & 88.62+8.06           \\ \cmidrule(l){2-7} 
                              & TD3BC+ Critic Defense & \textit{94.95+10.45} & \textit{93.35+5.31}  & \textit{48.33+13.36} & \textit{85.51+6.13}  & 63.96+22.2           \\ \cmidrule(l){2-7} 
                              & TD3BC+ Actor Defense  & \textit{106.68+1.25} & \textit{104.46+0.67} & \textit{76.51+6.16}  & \textit{102.44+1.31} & 92.78+10.53          \\ \midrule
\multirow{3}{*}{Hopper hop}   & TD3BC                 & \textit{112.27+0.28} & \textit{111.75+1.09} & \textit{79.31+20.65} & \textit{98.13+3.69}  & 102.44+11.99         \\ \cmidrule(l){2-7} 
                              & TD3BC+ Critic Defense & \textit{112.15+0.13} & \textit{111.38+1.32} & \textit{99.2+5.75}   & \textit{97.76+7.05}  & 102.68+11.62         \\ \cmidrule(l){2-7} 
                              & TD3BC+ Actor Defense  & \textit{112.07+0.15} & \textit{111.96+0.27} & \textit{95.24+8.95}  & \textit{105.98+5.09} & 86.99+27.35          \\ \midrule
\multirow{3}{*}{Half Cheetah} & TD3BC                 & 102.7+2.61           & 62.43+3.87           & 43.75+6.76           & 32.52+2.59           & 26.18+11.37          \\ \cmidrule(l){2-7} 
                              & TD3BC+ Critic Defense & 101.72+2.01          & 59.84+1.88           & 50.19+7.12           & 34.25+2.38           & 27.06+13.2           \\ \cmidrule(l){2-7} 
                              & TD3BC+ Actor Defense  & 97.3+3.11            & 60.45+5.69           & 44.78+11.6           & 36.55+1.81           & 24.11+10.78          \\ \midrule
\multirow{3}{*}{MEAN}         & TD3BC                 & 106.55+1.5           & 93.19+2.13           & 63.44+12.06          & 76.98+3.64           & 72.42+10.47          \\ \cmidrule(l){2-7} 
                              & TD3BC+ Critic Defense & 102.94+4.19          & 88.19+2.84           & 65.91+8.75           & 72.5+5.19            & 64.57+15.68          \\ \cmidrule(l){2-7} 
                              & TD3BC+ Actor Defense  & 105.35+1.5           & 92.29+2.21           & 72.18+8.9            & 81.66+2.74           & 67.96+16.22          \\ \bottomrule
\end{tabular}
\caption{The effectiveness of adversarial attacks the baseline TD3-BC and TD3-BC with defenses on Expert dataset.}
\label{tab:expert_performance}
\end{table*}